\documentclass{article}
\usepackage{multicol}
\usepackage{microtype}
\usepackage{graphicx}
\usepackage{subfigure}
\usepackage{booktabs} 

\usepackage{hyperref}
\usepackage{float}
\usepackage{algorithm}
\usepackage{algorithmic}

\usepackage{amsmath}
\usepackage{amssymb}
\usepackage{mathtools}
\usepackage{amsthm}

\usepackage{bbm}
\usepackage{amsfonts}       
\usepackage{amsthm} 

\theoremstyle{plain}
\newtheorem{theorem}{Theorem}[section]
\newtheorem{proposition}[theorem]{Proposition}

\theoremstyle{definition}
\newtheorem{definition}[theorem]{Definition}

\theoremstyle{remark}

\usepackage[textsize=tiny]{todonotes}

\newcommand{\RR}{\ensuremath{\mathbb{R}}}

\newcommand{\HH}{\ensuremath{\mathbb{R}^d}}
\newcommand{\LL}{\ensuremath{\mathbb{R}^k}}

\newcommand{\calB}{\mathcal{B}}

\newcommand{\calN}{\mathcal{N}}

\newcommand{\RRT}{\ensuremath{\mathbb{R}^T}}

\newcommand{\normone}[1]{{\|{#1}\|}_{1}}
\newcommand{\normtwo}[1]{{\|{#1}\|}_{2}}

\newcommand{\normQ}[1]{{\|{#1}\|}_{q}}
\newcommand{\normOneInf}[1]{{\|{#1}\|}_{1,\infty}}
\newcommand{\norminf}[1]{{\|{#1}\|}_{\infty}}
\providecommand{\keywords}[1]
{
  \small	
  \textbf{\textit{Keywords---}} #1
}

\begin{document}

\title{Multi-level projection with exponential parallel speedup; 
           Application to sparse  neural networks }

\author{Guillaume Perez and Michel Barlaud\\
I3S Laboratory, University of Cote d'Azur, France}

\vskip 0.3in
\maketitle

\begin{abstract}
The $\ell_{1,\infty}$ norm is an efficient structured projection but the complexity of the best algorithm is unfortunately $\mathcal{O}\big(n m  \log(n m)\big)$ for a matrix in $\RR^{n\times m}$.
In this paper, we propose a new bi-level projection method for which  
we show that the time complexity for the $\ell_{1,\infty}$ norm is only $\mathcal{O}\big(n  m \big)$ for a matrix in $\RR^{n\times m}$, and $\mathcal{O}\big(n + m \big)$ with full parallel power.
We generalize our method to tensors and we propose a new multi-level projection, having an induced decomposition that yields a linear parallel speedup up to an exponential speedup factor, resulting in a time complexity lower-bounded by the sum of the dimensions, instead of the product of the dimensions. 
we provide a large base of implementation of our framework for bi-level and tri-level (matrices and tensors) for various norms and provides also the parallel implementation.
Experiments show that our projection is $2$ times faster than the actual fastest Euclidean algorithms while providing same accuracy and better sparsity in neural networks applications.
\end{abstract}

\section{Introduction}
\subsection{Motivations and related methods}
Sparsity requirement appears in many machine learning applications,
such as the identification of biomarkers in biology \cite{abeel2009robust,he2010stable} 
or the recovery of sparse signals in compressed sensing \cite{donoho2006compressed,wright09}. 
It is well known that the impressive performance of neural networks is achieved at the cost of a high-processing complexity and large memory requirement.
Recently, advances in sparse recovery and deep learning have shown that training neural networks with 
sparse weights not only improves the processing time,
but most importantly improves the robustness and test accuracy of the learned models \cite{Alv2016,Han2015-2,Tar2018,Gom2019,Osw2016}\cite{dropout1,dropout2}. 
Regularizing techniques have been proposed to sparsify neural networks, such as the popular \textit{LASSO} method \cite{tRS,sparse}. The LASSO considers the  $\ell_1$ norm as regularization.
Note that projecting onto the $\ell_1$ norm ball is of linear-time complexity \cite{condat,Perez19}. 
Unfortunately, these methods generally produce sparse weight matrices, but this sparsity is not structured  and thus is not computationally processing efficient. 

Let's consider again the sparsity requirement. In neural networks processing, a lot of
hardware can compute multiply-add operations in a single
instruction. We therefore require low  MACCs (multiply-accumulate
operations) as computational cost (one multiplication
and one addition are counted as a single instruction).
Thus a structured sparsity is required (i.e. a sparsity able to set a whole set of columns to zero). 
To address this issue, extension of the $\ell_1$ to $\ell_{1,q}$ norms with $q \in 1,2, \infty$
promotes group sparsity, i.e., the solution is not only
sparse but group variables can be removed simultaneously.
 Group-LASSO originally proposed in \cite{Yua}, was used  in order to sparsify neural networks without loss of performance \cite{Hua2018,Yoo2017,Sca2017}. Unfortunately, the classical Group-LASSO algorithm is based on Block coordinate descent \cite{BCD,Fastlasso} which requires high computational cost.
Note that he $\ell_{1,1}$ constraint was also used in order to  sparsify  convolutionnal neural networks \cite{BG20}.\\
The $\ell_{1,\infty}$ norm is 
of particular interest because it is able to set a whole set of columns to zero, 
instead of spreading zeros as done by the $\ell_1$ norm.
This makes it particularly interesting for reducing computational cost.
Many projection algorithms were proposed
\cite{quattoni2009efficient,Rakoto,bejar2021fastest}.
However complexity of these algorithms remains an issue. The worst-case time complexity of this algorithm is $\mathcal{O}\big(nm*\log(nm)\big)$ for a matrix in $\RR^{n\times m}$.
This complexity is an issue, and to the best of our knowledge, no current publication reports the use of the $\ell_{1,\infty}$ projection for sparsifying large neural networks.

\subsection{Contributions}
In this paper we focus on the computational issue for the $\ell_{1,\infty}$ norm.
First, in order to cope with this issue, we propose a new bi-level projection method. 
The main motivation for this work is the direct independent splitting made by the bi-level optimization which take into account the structured sparsity requirement.
Second, we generalize the bi-level projection to multi-level, and shows its theoretical guaranties, such as an exponential parallel speedup.
The paper is organized as follows. First we present our general bi-level projection framework using our splitting approach \ref{new}. 
Then,  we provide in section \ref{bi-level-1Infty} the application to the bi-level $\ell_{1,\infty}$ projection. 
In section \ref{extend},  we apply our  bi-level framework to well known constraints such as $\ell_{1,1}$ constraint  and to any combination of p, q  norm. 
We extend our bi-level projection method to tensors using multi-level projections in section \ref{multi}. 
In Section \ref{Exp}, we finally
compare different projection methods experimentally.
Our experimental section is split in two parts.
First, we provide an empirical analysis of the projection algorithms onto the  bi-level projection $\ell_{1,\infty}$ ball.
This section shows the benefit of the proposed method, especially in the context of sparsity. 
Second, we apply our framework to the diagnosis (or classification) using a neural network (Autoencoder)  on a synthetic dataset and a biological dataset.

\section{Definitions and Notations}
In this paper we use the following notations:lowercase Greek symbol for scalars, scalar  i,j,c,m,n are indices of vectors and matrices, lowercase for vectors, capital for matrices, and calligraphic for tensors. 
For every $p,q \in 1,2, \cdot, \infty$, the 
$\ell_{p,q}$
norm of a real matrix 
$X = [x_1\ \cdots\ x_m] \in \RR^{n\times m}$ with columns $x_j$ and elements $X_{i,j}$ 
is given by:
\begin{equation}
  \|X\|_{p,q} := \left(\sum_{j=1}^m \|x_j\|_q^p\right)^{\frac{1}{p}},
\end{equation}
where the $\ell_q$ norm of the column vector $x_j\in\mathbb{R}^n$ is:
\begin{equation} 
  \|x_j\|_{q} := \left(\sum_{i=1}^n |X_{i,j}|^q\right)^{\frac{1}{q}}.
\end{equation}

Given a real value $\eta$, we denote by $\calB_\eta^{p}$ the ball of radius $\eta$ for the norm $p$:
\begin{equation}
\calB_\eta^{p}=\left\{X\in\RR^{n}| \;\;\|X\|_{p}\leq \eta\right\},
\end{equation}
Using the definition of the ball, the euclidean projection is defined by:
\begin{equation}\label{eq:projBasis}
P^{p,q}_\eta(Y)= \arg \min\limits_{X \in \calB_\eta^{p,q}} \normtwo{X-Y}
\end{equation}
In the general case, $p$ and $q$ are set and dedicated algorithms are used to process the projection.

\section{A new bi-level projection formulation}
\label{new}
Processing the projection of matrices or other mathematical objects, such as tensors, is becoming harder and harder as dedicated algorithms satisfying equation~(\ref{eq:projBasis}) must be defined for each possible combination of $p$ and $q$.
In this section we relax equation~(\ref{eq:projBasis}) and to reformulate the projection as a bi-level problem.
In this reformulated problem, the $\ell_{p,q}$ projection will be split into an aggregation using the $q$ norm followed by a simpler and well defined projection onto the $p$ norm of the aggregated vector.
Let $v_{q} = (\normQ{y_{1}},\dots,\normQ{y_{m}})$ be the raw vector composed of the $q$ norms of the columns of the matrix Y.\\ 
Given a real positive value $\eta$.
The $\ell_{p,q}$ bi-level projection optimization problem is defined by:
\begin{equation}
\begin{aligned}
BP^{p,q}_\eta(Y)=\{X|\forall j, x_j = \arg\min\limits_{x_j \in \calB_{\hat{u}_j}^{q}} \normtwo{x_j-y_j}\\
\text{such that}\quad 
\hat{u} \in \arg \min\limits_{u \in \calB_\eta^{p}} \normtwo{u-v_{q}}\}
\end{aligned}
\end{equation}

This problem is composed of two simpler problems. 
The first one, the most inner one is:
\begin{equation}
\hat{u} \in \arg \min\limits_{u \in \calB_\eta^{p}} \normtwo{u-v_{q}}
\end{equation}
Once the columns of the matrix have been aggregated to a vector $v_{q} $ using the $q$ norm ,
the problem becomes a usual $\ell_{p}$ ball projection problem.
Such problems are solved by definition by:
\begin{equation}
    u \gets P^{p}_\eta((\normQ{y_{1}},\dots,\normQ{y_{m}}))
\end{equation}
It is interesting to note that for some values of $p$, projection algorithms have been defined already. 
For example, linear algorithms exist for the $\ell_{1}$ and its weighted version $\ell_{w1}$, the $\ell_{2}$ and $\ell_{\infty}$. 
In addition all of them are strongly convex, which make the solution unique.
Note that it is not necessarily the case for any $p$.

Then, the second part of the bi-level optimization problem, once vector $\hat{u}$ is known, for each column $j$ of the original matrix $x_j $ an independent optimization problem is defined:
\begin{equation}
x_j = \min\limits_{x_j \in \calB_{\hat{u}_j}^{q}} \normtwo{x_j-y_j}
\end{equation}

Each of column  $x_j$ is optimally solved by definition of the projection on the $\ell_{q}$ ball:
\begin{equation}
x_j \gets P^{q}_{u_j}(y_j) \forall j \in 1,\dots,m
\end{equation}
The following algorithm is a possible implementation.

\begin{algorithm}[h]
   \caption{Bi-level $\ell_{p,q}$ projection ($BP^{p,q}_\eta(Y)$).}
   \label{algo:pqgeneral}
\begin{algorithmic}  
\STATE \textbf{Input:} $Y,\eta$
\STATE{$u \gets P^{p}_\eta((\normQ{y_{1}},\normQ{y_{j}},\dots,\normQ{y_{m}}))$}  
\FOR{$j \in [1,\dots,m]$}
  \STATE $x_j \gets P^{q}_{u_j}(y_j)$ 
\ENDFOR
\STATE \textbf{Output:} $X$
\end{algorithmic}
\end{algorithm}

Here again, the projection onto the $\ell_{q}$ should be known and well defined.
It is important to remark that even if the projection is not the closest from a Euclidean point of view, the resulting matrix satisfies the constraint (i.e. $X \in \calB_\eta^{p,q}$).\\
In the following sections provide our bi-level projections and compare with the existing projections in the literature.
It is important to note that we focused on norms providing structured sparsity (removing columns), namely bi-level  $\ell_{1,\infty}$, bi-level  $\ell_{1,1}$ and bi-level  $\ell_{1,2}$.
Yet, this is not exhaustive as more norms using different values for $p$ and $q$ can be defined.

\section{Bi-level  $\ell_{1,\infty}$ projection}

\label{bi-level-1Infty}
\subsection{A new bi-level projection}

The $\ell_{1,\infty}$ ball projection has gained a lot of interest in the recent years. The main reason being its efficiency to enforce sparsity in the weights of neural networks, while keeping high accuracy.\cite{quattoni2009efficient,chau2019efficient,chu2020semismooth,bejar2021fastest} and the classical approach is given as follows.\\
Let $Y \in \RR^{n \times m}$ be a real matrix of dimensions $m\geq 1$, $n\geq 1$, 
with elements 
$Y_{i,j}$, $i=1,\ldots,n$, $j=1,\ldots,m$. 
The $\ell_{1,\infty}$ norm of $Y $ is
\begin{equation}
    \normOneInf{Y} := \sum_{j=1}^m \max_{i=1,\ldots,n} |Y_{i,j}|.
\end{equation}
Given a radius $\eta \geq 0$, the goal is to project $Y$ onto the   $\ell_{1,\infty}$ norm ball of radius $\eta$, denoted by 
\begin{equation}
\calB^\eta_{1,\infty}:=\left\{X\in\RR^{n \times m}\ : \ \normOneInf{X}\leq \eta\right\}.
\end{equation}
The projection $P_{\calB^\eta_{1,\infty}}$ onto $\calB^\eta_{1,\infty}$ is given by:
\begin{equation}
P_{\calB^\eta_{1,\infty}} = {arg \min \limits_{X \in \calB^\eta_{1,\infty}}   \frac{1}{2} \|X-Y\|_\mathrm{F}^2} 
\end{equation}
where $\|\cdot\|_\mathrm{F}=\|\cdot\|_{2,2}$ is the Frobenius norm. \\

In this paper we propose the following alternative new bi-level method.
Recall that ${v_\infty} = (\norminf{y_{1}},\dots,\norminf{y_{m}})$ is the vector composed of the infinity norms of the columns of matrix $Y$.
Let the infinity norm projection $P^{\infty}_{u_i}(y) = (\min(y_i,u_i),\forall y_i \in y)$.
The bi-level $\ell_{1,\infty}$ projection optimization problem is defined by:
\begin{equation}
\begin{aligned}
BP^{1,\infty}_\eta(Y)=\{X|\forall j, x_j =\arg \min\limits_{X_j \in \calB_{\hat{u}_j}^{\infty}} \normtwo{x_j-y_j}\\
\text{such that}\quad 
\hat{u} \in \arg \min\limits_{u \in \calB_\eta^{1}} \normtwo{u-v_{\infty}}\}
\end{aligned}
\end{equation}
Algorithm~\ref{algo:linfbiproj} is a possible implementation.
It is important to remark that usual bi-level optimization requires many iterations \cite{bilevel2,bilevel3}
while our model reaches the optimum in one iteration.

\begin{algorithm}[h]
   \caption{Bi-level $\ell_{1,\infty}$ projection ($BP^{1,\infty}_\eta(Y)$).}\label{algo:linfbiproj}
\begin{algorithmic}  
\STATE \textbf{Input:} $Y,\eta$
\STATE{$u \gets P^{1}_\eta((\norminf{y_{1}},\norminf{y_{j}},\dots,\norminf{y_{m}}))$}  
\FOR{$j \in [1,\dots,m]$}
  \STATE $x_j \gets P^{\infty}_{u_j}(y_j)$ 
\ENDFOR
\STATE \textbf{Output:} $X$
\end{algorithmic}
\end{algorithm}

\subsection{Computational complexity}
The best computational complexity of the projection of a matrix in $\RR^{nm}$ onto the $\ell_{1,\infty}$ ball is usually $O(nm\log(nm))$ \cite{quattoni2009efficient,bejar2021fastest}.
The computational complexity of the bi-level projection here is $O(nm)$.
Indeed, consider Algorithm \ref{algo:linfbiproj}.
Step 1) complexity is $O(nm)$, step 2) complexity is $0(n)$ \cite{condat,Perez-2023}, step 3) complexity is $0(n)$, and step 4 complexity is $O(nm)$.
Moreover, the bi-level projection explicit independent processing.
While the time complexity without any parallel processing will remain the same,
the time complexity with a full parallel power is only $O(n + m)$ as steps 1) and 4) can be run in a parallel.

\begin{table*}[t]
    \centering  
    \begin{tabular}{|l|c|c|c|c|c|c|c|c|}
    \hline
       Synthetic  & bi-level $\ell_{1,2}$  & $\ell_{1,1}$  & bi-level $\ell_{1,1}$ & $\ell_{1,\infty}$  & bi-level $\ell_{1,\infty}$  \\
       \hline
        Complexity $ $ & $O(mn)$ & $O(mn)$ & $O(mn)$ & $O(nm\log(nm))$ & $O(nm)$ \\
        \hline
        LP Complexity $ $ & $O(m+n)$ & $O(mn)$ & $O(m+n)$ & $O(nm\log(nm))$  & $O(n+m)$\\
        \hline
    \end{tabular}
    \caption{\textbf{Theoretical complexity} Complexity assuming the  $\ell_{1}$ and $\ell_{2}$ projection costs are linear \cite{Perez-2023} and  complexity of $\ell_{1,\infty}$ is $O(mn\log(nm))$ \cite{quattoni2009efficient}. LP (i.e. longest path) complexity is the complexity of the longest path in the computational graph of the bi-level projection. This is the complexity limit of a generic parallel implementation.}
    \label{Complexity}
\end{table*}

\subsection{Extension to other norms }
\label{extend}
\subsubsection{Extension to the $\ell_{1,1}$  bi-level projection  }
The $\ell_{1}$ ball is famous for being robust and yielding sparsity.
Nevertheless, its extension to matrix, the $\ell_{1,1}$ does not yield structured sparsity.
Let ${v_1} = (\normone{y_{1}},\dots,\normone{y_{m}})$ the raw vector composed of the $\ell_1$ norm of the columns of matrix $Y$.
We propose to define the $\ell_{1,1}$ bi-level optimization problem:
\begin{equation}
\begin{aligned}
BP^{1,1}_\eta(Y)=\{X|\forall j, x_j = \arg \min\limits_{x_j \in \calB_{\hat{u}_j}^{1}} \normtwo{x_j-y_j}\\
\text{such that}\quad 
\hat{u} \in \arg \min\limits_{u \in \calB_\eta^{1}} \normtwo{u-v_{1}}\}
\end{aligned}
\end{equation}

This bi-level $\ell_{1,1}$ projection yields structured sparsity. 
A possible implementation of the bi-level $\ell_{1,1}$ is given in Algorithm~\ref{algo:li1iproj}.
This bi-level $\ell_{1,1}$ projection has the same advantage as the 
bi-level $\ell_{1,\infty}$ which is the induced parallel decomposition leading to a 
time complexity with a full parallel power of $O(n + m)$.

\begin{algorithm}[h]
   \caption{Bi-level $\ell_{1,1}$ projection. ($BP^{1,1}_\eta(Y)$)}\label{algo:li1iproj}
\begin{algorithmic}  
\STATE \textbf{Input:} $Y,\eta$
\STATE{$u \gets P^{1}_\eta((\normone{y_{1}},\dots,\normone{y_{m}}))$} 
\FOR{$j \in [1,\dots,m]$}
  \STATE $x_j \gets P_{u_j}^{1}(y_j)$
\ENDFOR
\STATE \textbf{Output:} $X$
\end{algorithmic}
\end{algorithm}

\subsubsection{Extension to bi-level $\ell_{1,2}$ projection}
As using $p=1$ yields sparsity in our experiments, we propose to also consider the bi-level $\ell_{1,2}$ projection.
Let ${v_2} = (\normone{y_{1}},\dots,\normone{y_{m}})$ the raw vector composed of the $\ell_2$ norm of the columns of matrix $Y$.
The $\ell_{1,2}$ bi-level optimization problem criterion is:
\begin{equation}
\begin{aligned}
BP^{1,2}_\eta(Y)=\{X|\forall j, x_j = \arg \min\limits_{x_j \in \calB_{\hat{u}_i}^{2}} \normtwo{x_j-y_j}\\
\text{such that}\quad 
\hat{u} \in \arg \min\limits_{u \in \calB_\eta^{1}} \normtwo{u-v_{2}}\}
\end{aligned}
\end{equation}

The bi-level projection algorithms for $\ell_{1,2}$ is given by algorithm~\ref{algo:li12proj}.
\begin{algorithm}[h]
   \caption{Bi-level $\ell_{1,2}$ projection. ($BP^{1,2}_\eta(Y)$)}\label{algo:li12proj}
\begin{algorithmic}  
\STATE \textbf{Input:} $Y,\eta$
\STATE{$u \gets P^{1}_\eta((\normtwo{y_{1}},\dots,\normtwo{y_{m}}))$} 
\FOR{$j \in [1,\dots,m]$}
  \STATE $x_j \gets P^{2}_{u_j}(y_j) \forall j \in 1,\dots,m$
\ENDFOR
\STATE \textbf{Output:} $X$
\end{algorithmic}
\end{algorithm}
Note that the bi-level $\ell_{1,2}$ projection and it's relationship to other existing norms is out of the scope of this paper.
Yet we propose to use this algorithm as a comparison in our experimental section.

\section{Tensor Generalization }
\label{multi}
\subsection{Tri-level Projection Algorithms}
Matrices are not the only mathematical objects that are used in neural networks.
Images, for examples, are often represented as order 3 tensors in $\RR^{c,n,m}$ with $c$ representing the channels of the image.
Moreover, the current development of deep-learning framework for image compression are leaning toward tensor efficient regularization methods \cite{Twitter,Mentzer},
and not only vectors or matrices methods.
For example, the new image compression standard JPEG AI \cite{JPEGAI} uses tensor representation of images in the latent space of an autoencoder.
That is the reason why we propose to generalize the bi-level projection, first to tri-level, then to multi-level.
We consider tensors as multi-dimensional arrays with subscripts indexing.
First, consider order 3 tensors of the form $\mathcal{Y} \in \RR^{c,n,m}$.
For example, $\mathcal{Y} \in \RR^{3,m,n}$ is used to represent an image of width $m \times n$ and having 3 channels.
We define by $\forall i$ $\in <c,n,m>$ all the tuples of the Cartesian product 
of the indices. 
$i$ is a multi-dimensional index, not a single value.
For example $\forall i$ $\in <3,256,256> $ $\equiv \forall i \in ((1,1,1),(1,1,2),\dots,(3,256,256))$.
Given a tensor $\mathcal{Y} \in \RR^{c,n,m}$, we denote by $V_{q} = (\normQ{\mathcal{Y}_{1,1}},\dots,\normQ{\mathcal{Y}_{n,m}})$ 
the matrix of aggregation of the channels by the norm $q$.

\begin{definition}
Given a tensor $\mathcal{Y} \in \RR^{c,n,m}$ and a radius $\eta$, the tri-level $\ell_{1,\infty,\infty}$ projection optimization problem is equal to:
\begin{eqnarray}
\begin{aligned}
TP^{1,\infty,\infty}_\eta(\mathcal{Y})=\{\mathcal{X}|\forall i \in <n,m>, \mathcal{X}_{i} = \\
\arg \min\limits_{ \mathcal{X}_{i} \in \calB_{\hat{U}_{i}}^{\infty}} \normtwo{\mathcal{X}_{i}-\mathcal{Y}_{i}}. \\
\quad \text{such that}\quad 
\hat{U} \in BP^{1,\infty}_\eta(V_\infty)\}
\end{aligned}
\end{eqnarray}
\end{definition}

This tri-level projection first uses the $\ell_{\infty}$ norm as a multi-channel aggregator.
Then, for each column of the resulting matrix, it aggregates again using the $\ell_{\infty}$ norm.
Finally, the resulting vector is projected onto the to the $\ell_{1}$ ball.
A possible implementation is given in Algorithm~\ref{algo:linfimageproj},
where $BP^{1,\infty}_\eta(V_{\infty})$ is given in Algorithm~\ref{algo:linfbiproj}.
First, line \ref{algo:linfimageproj:aggreg} shows the aggregation of the tensor into a vector. 
The infinity norm aggregation is applied to the channels, 
then the previously defined bi-level $\ell_{1,\infty}$ is applied and stored in $U^{2}$.
Finally, for each couple of coordinate $(i,j)$, 
the resulting channels is equal to the original channels at the coordinate projected onto the 
$\ell_{\infty}$ ball of radius $U_{i,j}$.
An iterative implementation of the algorithm is given in Algorithm~\ref{algo:linfimageprojIter} in appendix.

\begin{algorithm}[h]
   \caption{Tri-level projection $\ell_{1,\infty,\infty}$. ($MP^{1,\infty,\infty}_\eta(\mathcal{Y})$)}\label{algo:linfimageproj}
\begin{algorithmic}[1]
\STATE \textbf{Input:} $\mathcal{Y} \in \RR^{c,n,m},\eta$ 
\STATE{$U \gets BP^{1,\infty}_\eta(V_{\infty})$} \COMMENT{Algorithm~\ref{algo:linfbiproj}} \label{algo:linfimageproj:aggreg}
\FOR{$ i \in <n,m>$}
    \STATE $x_i \gets P^{\infty}_{U_i}(y_{i})$   
\ENDFOR
\STATE \textbf{Output:} $\mathcal{X}$
\end{algorithmic}
\end{algorithm}

\subsection{Multi-level Projection Algorithms }
As shown in the previous section,
the tri-level projection yields a recursive pattern that is generalized in this section.

We consider tensors as multi-dimensional arrays with subscripts indexing.
Let $T=(d_1,\dots,d_r)$ a list of $r$ dimensions, usually called shape.
Let $|s|$ and $s_{i:j}$ respectively denote the cardinality and sub-list operator. 
Let $\mathcal{Y} \in \RRT$ denotes a tensor of order $|T|$.
For example if $T=(c,n,m)$, then $\mathcal{Y} \in \RRT$ is an order 3 tensor.

Let $\nu \in \calN^{|T|}$ a list of norms design of an order $|T|$ tensor.
Let each norm $\nu^i \in \nu$ be of the form $\nu^i=(\nu^i_1,\dots,\nu^i_j)$.
Let $\forall i \in <T_{k:l}>$ denote all the tuples of the Cartesian product 
of the indices from the sub-list of dimensions $T_{k:l}$. 
This is an extension to the the order 3 definition of the indexes enumeration.

Given $\mathcal{Y} \in \RRT$ let $\mathcal{V}_{\nu^i} \in \RR^{T_{1:|T|-|\nu^i|}}$ 
be the aggregation of $\mathcal{Y}$ using the $\nu^i$ norm.

\begin{definition}
Given a tensor $\mathcal{Y} \in \RRT$, a list of norms $\nu$, and a radius $\eta$, the multi-level projection optimization problem is equal to:
\begin{eqnarray}
\begin{aligned}
MP^{\nu}_\eta(\mathcal{Y})=\{\mathcal{X}|\forall i \in <T_{|\nu_1|:|T|}>, \mathcal{X}_{i} = \\
\arg \min\limits_{ \mathcal{X}_{i} \in \calB_{\hat{\mathcal{U}}_{t}}^{\nu_1}} \normtwo{\mathcal{X}_{i}-\mathcal{Y}_{i}}. \\
\quad \text{such that}\quad 
\hat{\mathcal{U}} \in MP^{\nu_{2:|\nu|}}_\eta(\mathcal{V}_{\nu_1})\}
\end{aligned}
\end{eqnarray}
and $MP^{\nu}_\eta(\mathcal{Y})=P^{\nu_1}_\eta(\mathcal{Y})$ if $|\nu|=1$.
\end{definition}
For example $\nu=((1,\infty))$ is be the usual $\ell_{1,\infty}$ norm
while $\nu=((\infty),(1))$  is be the bi-level $\ell_{1,\infty}$ norm.
Finally, if $\nu=((\infty),(\infty),(1))$ it is the tri-level defined in the previous section.
\begin{proposition}
    The multi-level projection is a generalization of the usual projection.
\end{proposition}
Proof is in appendix.
In the general case, the computational worst-case time complexity of a projection is lower-bounded by the product of the dimensions of $\mathcal{Y}$ given by $O(\prod_{d\in T_r} d)$.
For example, the best known computational complexity of the projection of a matrix in $\RR^{n\times m}$ onto the $\ell_{1,\infty}$ ball is usually $O(nm\log(nm))$ \cite{quattoni2009efficient,bejar2021fastest}.
Yet sometimes, for some norms, a parallel version can be defined, or at least with independent sub-parts.
Multi-level projection aims at reducing this complexity.

Algorithm~\ref{algo:linfmultiproj} is a possible implementation of the multi-level projection.
At the first line, the recursive call extracts the multi-level projection of the $\nu_1$-aggregated tensor onto
the list of norms $\nu_{2:|\nu|}$. 
This list consists of all the norms except the first one.
For example, for the tri-level $\nu=((\infty),(\infty),(1))$, the first line 
is the bi-level projection of the tensor aggregated using the $\infty$ norm onto the list of norm $((\infty),(1))$.
Then each tuple of indices $t$ in $<T_{|\nu_1|:|T|}>$ if considered.
It is the set of indices of all the dimensions except the $|\nu_1|$ first ones. 
Consider again the tri-level example, the $t$ was among the set of indices $<n,m>$.
For each of these tuples $t$, the local projection onto $\nu_1$ of the sub-part $\mathcal{Y}_t$ of the tensor $\mathcal{Y}$ is processed
and stored in the sub-part $\mathcal{X}_t$ the tensor $\mathcal{X} \in \RRT$.
In the end, $\mathcal{X}$ contained the multi-level projection of $\mathcal{Y}$.

\begin{proposition}
Using infinite parallel processing power, the lower-bound worst-case time complexity of the multi-level projection is reduced from $O(\prod_{d\in T_r} d)$ to $O(\sum_{d\in T_r} d)$, resulting in an exponential speedup.
\end{proposition}
Proof is given in appendix.
This implies that with infinite parallel processing power, the time complexity of the aggregations of all the recursive calls is $O(\sum_i f_i(|\nu_i|))$.

\begin{algorithm}[ht]
   \caption{Multi-level projection ($MP^{\nu}_\eta(\mathcal{Y})$).}\label{algo:linfmultiproj}
\begin{algorithmic}[1] 
\STATE \textbf{Input:} $\mathcal{Y},\nu,\eta$
\STATE{$\mathcal{U} \gets MP^{\nu_{2:|\nu|}}_\eta(\mathcal{V}_{\nu_1})$} \label{algo:linfmultiproj:aggreg} 
\FOR{$ i \in <T_{|\nu_1|:|T|}>$}\label{algo:linfmultiproj:loopt} 
    \STATE $x_i \gets P^{\nu_1}_{\mathcal{U}_i}(y_i)$ \label{algo:linfmultiproj:project} 
\ENDFOR
\STATE \textbf{Output:} $\mathcal{X}$
\end{algorithmic}
\end{algorithm}

\section{Experimental results} 
\label{Exp}
\subsection{Benchmark times using Pytorch C++ extension using a Macbook Labtop with with a i9 processor; Comparison with the best actual projection method }
This section presents experimental results of the projection operation alone.
The experiments were run on laptop with  a I9 processor having 32 GB of memory.
The state of the art on such is pretty large, starting with \cite{quattoni2009efficient} who proposed the first algorithm, the Newton-based
root-finding method and column elimination method \cite{chau2019efficient,bejar2021fastest}, and the recent paper of \textit{Chu et. al.} \cite{chu2020semismooth} which outperforms all of the other state-of-the-art methods.
We compare our bi-level method against the best actual algorithm proposed by \textit{Chu et. al.}
which uses a semi-smooth Newton algorithm for the projection. 
The Pytorch C++ extension implementation used is the one generously provided by the authors.
All other methods usually take order of magnitude more times, 
hence are not present in most of our figures.\\
The Pytorch  C++ implementation of our bi-level $\ell_{1,\infty}$ method is based on fast
$\ell_{1}$ projection algorithms of \cite{condat,Perez19} which are of linear complexity.
The code is available online\footnote{https://github.com/memo-p/projection}.\\

\begin{figure}[t]
    \centering
    \includegraphics[width=0.8\textwidth,height=4.2cm]{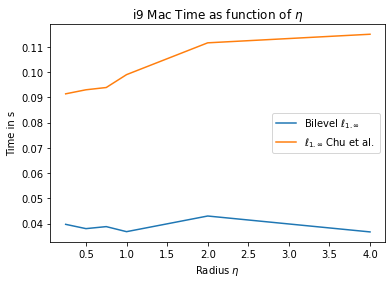}
    \caption{Comparison of our bi-level method with \textit{Chu et .al.}.  Processing time as a function of the radius (matrice fixed 1000x10000)}
    \label{fig:bi-levelcppChu}
\end{figure}
Figure~\ref{fig:bi-levelcppChu} shows the running time as a function of the radius. 
The size of the matrices is 1000x10000, values between 0 and 1 uniformly sampled and the radius are in $[0.25, 4]$.
As we can see, the running time of our bi-level method   is at least 2.5 times faster that the actual fastest method \textit{Chu et al.}. 
Note that this behaviour is the same for all the $\ell_{1,\infty}$ projection algorithms we tested.
The running time of the bi-level one is almost not impacted by the sparsity, which make it more stable in general.

\begin{figure}[t]
    \centering
    \includegraphics[width=0.8\textwidth,height=4.2cm]{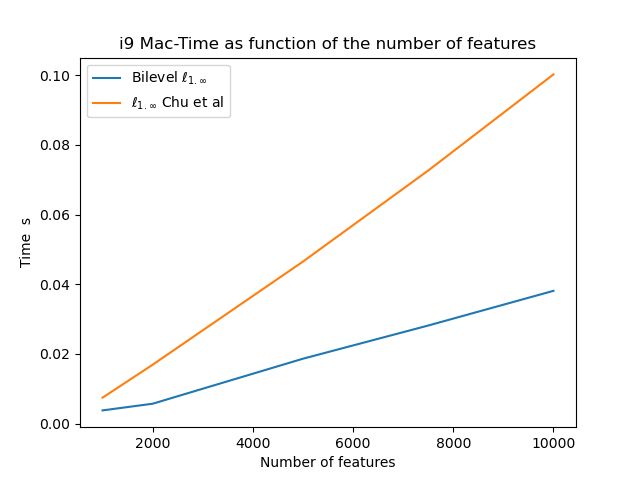}
    \caption{Processing time using C++: our bi-level projection method versus \textit{Chu et. al.} method.
     m=1000 and  $\eta=1$ fixed . }
    \label{fig:Chu versus bilevel}
\end{figure}

Figure~\ref{fig:Chu versus bilevel} shows the running time as a function of the matrix size. 
Here the radius has been fixed to $\eta=1.$.
As we can see, the running time of our bilevel method is at least 2.5 faster that  that the actual fastest method, and this factor remains the same even when increasing the matrix size both in number of columns and rows.
In conclusion to this comparison, using the bi-level $\ell_{1,\infty}$ is faster in our experiments and smoother in the selection of a radius. Note that pytorch c++ extension is 20 times faster than the standard pytorch implementation. 
\begin{figure}[t]
    \centering
    \includegraphics[width=0.8\textwidth,height=4.2cm]{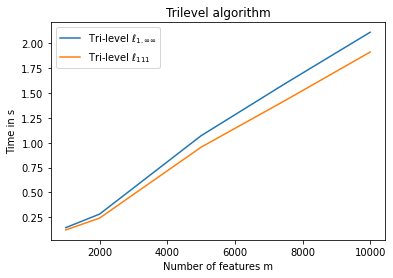}    
    \caption{Processing time using C++: our Tri-level projection method d= 32 and n=1000 fixed. }
    \label{trilevel}
\end{figure}\\
Figure~\ref{trilevel} shows  the running times of our Tri-level algorithm as a function of the tensor size with dimension m for two projections $\ell_{1,1,1} $and  $\ell_{1,\infty,\infty}$.
We can see that the running times of both projections are similar and  grow linearly with the increase of the dimension m. Note that we have the same behavior for the other dimensions of the tensor. 

\subsection{Benchmark using C++ parallel implementation}
One of the arguments of using the bi-level version of projection instead of the usual one is not only that it can be faster and easier to implement, but also that a native parallel decomposition of the work can be extracted from the computation tree.
We implemented the parallel version of the bi-level  $\ell_{1,\infty}$ using a basic Thread-pool implementation using native future of C++.
These experiments were run on a 12-Core Processor an \textit{AMD Ryzen 9 5900X 12-Core Processor 3.70 GHz} desktop machine having 32 GB of memory. \\, thus we used 12 as a maximum number of workers.
In this run, the compiler optimization have been deactivated. 
The reason is that once activated, the best speed factor is 2, 
larger instances are required to see better speedup, 
and the time is spent moving memory around and saturating the few memory channels.
Yet, future work involving CPU and GPU expert might solve these issues.
\begin{figure}[t]
    \centering
    \includegraphics[width=0.8\textwidth,height=4.5cm]{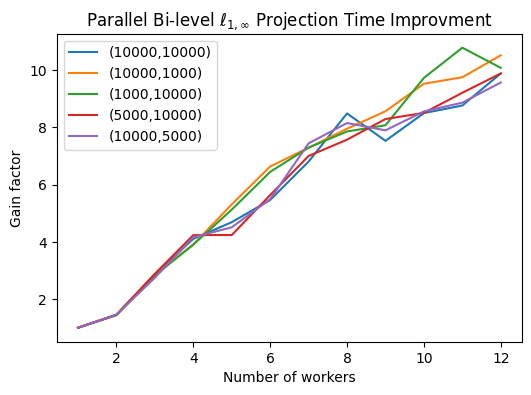}
    \caption{Gain factor of the native parallel workload decomposition for various matrix sizes.}
    \label{fig:parallel}
\end{figure}
As shown in Figure~\ref{fig:parallel}, the workload is easy to balance between workers by definition of the processing tree.
That is the reason why the gain factor grows linearly with the number of workers.
This implies that in our computer, our naive parallel implementation using CPU is between 20 to 30 times faster than the best current $\ell_{1,\infty}$ projection algorithm.

\subsection{Experimental results on classification using a supervised autoencoder neural network}
\subsubsection{Supervised Autoencoder (SAE) framework}
Autoencoders were introduced within the field of neural networks decades ago, their most efficient application  being dimensionality reduction \cite{deep,Twitter}. 
Autoencoders were used in application ranging from unsupervised deep-clustering \cite{VAE,semisupervised,Rochelle} to supervised learning adding a classification loss in order to improve classification performance \cite{LESAE,ICASSP}.
In this paper, we use the cross entropy as the added classification loss.\\
Let $X$ be the dataset in $\HH$, and $Y$ the labels in $\{0, \dots , k\}$, with $k$ the number of classes.
Let $Z \in \LL$ be the encoded latent vectors, $\widehat{X}$ $\in \HH $ the reconstructed data and $W$ the weights of the neural network.
Note that the dimension of the latent space $k$ corresponds to the number of classes.\\
The goal is to learn the network weights $W$ minimizing the total loss.
In order to sparsify the neural network,
we propose to use the different bi-level projection methods as a constraint to enforce sparsity in our model.
 The global criterion to minimize
 \begin{equation}
 \label{crit}
\underset{W}{\text{minimize}}  \quad \phi(X,Y) \quad\text{ subject to }  \quad BP^{1,\infty}(W) \leq \eta
\end{equation}
where $\phi(X,Y)=$ $\alpha \psi (X,\widehat{X})) +$ $\mathcal{H}(Y,Z)$. We use the Cross Entropy Loss$\mathcal{H}$  as the added classification loss 
and the robust Smooth $\ell_1$ (Huber) Loss \cite{Huber} as the reconstruction loss $\psi$.
Parameter $\alpha$ is a linear combination factor used to define the final loss.  
Note that the constrained approach \cite{BBCF} avoids computing the Lagrangian parameter with the "lasso path" in the case of a Lagrangian approach \cite{hrtzER,fht}, which is computationally costly \cite{myCA}. 
We compute the mask by using the various bilevel projection methods and we use the double descent algorithm \cite{Lottery,double} for minimizing the criterion \ref{crit}.
We implemented our SAE method using the PyTorch framework for the model, optimizer, schedulers and loss functions. 
We chose the ADAM optimizer \cite{Adam}, as the standard optimizer in PyTorch.
We used a symmetric linear fully connected network  with the encoder comprised of an input layer of $d$ neurons, one hidden layer followed by a ReLU  or SiLU activation function and a latent layer of dimension $k=2$.

\subsubsection{Experimental accuracy results}
We generate  artificial biological data to benchmark our bi level projection using  the $make\_classification$ utility from \textit{scikit-learn}. 
We generate $n=1,000$ samples  with a  number $m = 2000$ features because this is the typical range for biological data. We chose a low number of informative features ($ 64$ ) and a separability= $ 0.8$ realistically with  biological databases.\\
We provide the classical accuracy metric and the sparsity score in $ \%$: number of columns or features set to zero 

The biological \textbf{LUNG} dataset was provided by Mathe et al. \cite{Lung}. The goal of this experiment is to propose a diagnosis of the Lung cancer from urine samples. This dataset includes metabolomic data concerning urine samples from $10005$ samples: $469$ Non-Small Cell Lung Cancer (NSCLC) patients prior to treatment and $536$ control patients. Each sample is described by $m=2944$ metabolomic features. We apply to this metabolic dataset
 the classical log-transform for reducing heteroscedasticity.\\
Table \ref{SyntAccuracysparsity} shows accuracy classification. The baseline is an implementation that does not process any projection.
Compared to the baseline the SAE using the $\ell_{1,\infty}$ projection improves the accuracy by $7. \%$.\\
Table \ref{LungAccuracysparsity} shows that accuracy results of the bi-level  $\ell_{1,\infty}$ and classical $\ell_{1,\infty}$ are similar 
Again, compared to the baseline the SAE using the $\ell_{1,\infty}$ projection improves the accuracy by $3.5 \%$.

\begin{figure}[ht]
    \centering
    \includegraphics[width=0.49\textwidth,height=4.cm]{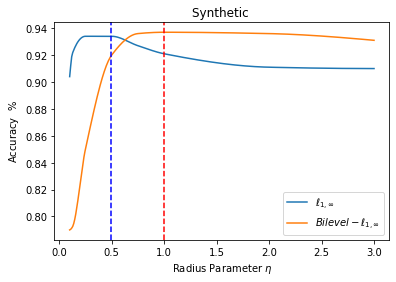}
    \includegraphics[width=0.49\textwidth,height=4.cm]{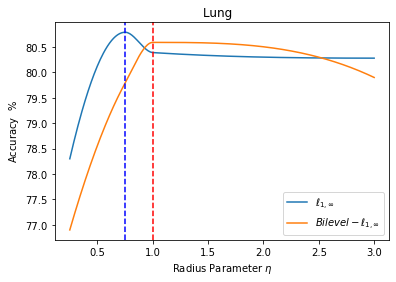}
    \caption{Accuracy  as a function of the radius parameter $\eta$; Left Synthetic data:, Right Lung dataset}
    \label{SyntRadius}
\end{figure}

From Table \ref{SyntAccuracysparsity} on synthetic dataset and Table \ref{LungAccuracysparsity} on Lung dataset it can be seen that the best accuracy is obtained for $\eta=0.75 $ for $\ell_{1,\infty}$ and for $\eta= 1. $ for the bilevel $\ell_{1,\infty}$ projection, maximum accuracy of both method are similar. 
However sparsity and computation  time are better for bilevel $\ell_{1,\infty}$ than for regular $\ell_{1,\infty}$.

\begin{table}[h!]
    \centering  
    \begin{tabular}{|c|c|c|c|c|}
    \hline
      Synthetic  & Baseline &$\ell_{1,\infty}$ & bi-level $\ell_{1,\infty}$   \\
      \hline  
      Best Radius & & 0.75 & 1.0 \\
      \hline   
        Accuracy  $ \%$ & 86.6$\pm 1.2$&94.4 $\pm 1.45$ & 94. $\pm 1.45 $ \\
        \hline   
        Sparsity $ \%$ &- &90.57 $\pm 0.021$ & 94.66 $\pm 0.02$ \\
        \hline       
    \end{tabular}
   \caption{\textbf{Synthetic } dataset.  SiLU activation, Accuracy versus sparsity  : comparison of  $\ell_{1,\infty}$ and  bi-level $\ell_{1,\infty}$.}
   \label{SyntAccuracysparsity}
\end{table}
\begin{table}[h!]
    \centering  
    \begin{tabular}{|c|c|c|c|c|}
    \hline
      Lung & Baseline &$\ell_{1,\infty}$ (Chu )& bi-level $\ell_{1,\infty}$   \\
       \hline 
       Best Radius & - & 0.75&  1.0 \\
      \hline   
        Accuracy  $ \%$ &77.12 $\pm 3$ &80.79 $\pm 0.4$ & 80.59 $\pm 0.43$ \\
        \hline   
        Sparsity  $ \%$ &- &95.63 $\pm 0.75$ & 95.83 $\pm 0.6$ \\
        \hline  
       
        \hline       
    \end{tabular}
    \caption{\textbf{Lung } dataset.  SiLU activation, Accuracy versus sparsity   : comparison of  $\ell_{1,\infty}$ $\eta= 0.75 $ and  bi-level $\ell_{1,\infty}$ $\eta= 1 $.}
    \label{LungAccuracysparsity}
\end{table}

So bilevel $\ell_{1,\infty}$ projection outperforms usual $\ell_{1,\infty}$ projection in terms of computation cost and sparsity.  

\section{Conclusion}

Although many projection algorithms
were proposed for the projection of the $\ell_{1,\infty}$  norm, complexity of these algorithms
remain an issue. The worst-case time complexity
of these algorithms is $\mathcal{O}\big(n \times m \times \log(n \times m)\big)$ for a matrix in $\RR^{n\times m}$. 
In order to cope with this complexity issue, we
have proposed a new bi-level (and multilevel) projection
method. The main motivation of our work is the
direct independent splitting made by the bi-level
optimization which takes into account the structured
sparsity requirement. We showed that the
theoretical computational cost of our new bi-level
method is only $\mathcal{O}\big(n \times m \big)$ for a matrix in $\RR^{n\times m}$.
Experiments on synthetic  data show that our
bi-level method is 2.5 times faster than the actual
fastest algorithm provided by \textit{Chu et. al.}. Moreover our bi-level $\ell_{1,\infty}$ projection  outperforms other bi-level projections $\ell_{1,1}$ and $\ell_{1,2}$.
Our parallel implementation  with Nw workers improves the computational time by a factor close to Nw. 
Note that our extension to multilevel projection can be applied for sparsifying large convolutional neural networks.

\bibliography{references}

\begin{thebibliography}{10}
\providecommand{\url}[1]{#1}
\csname url@samestyle\endcsname
\providecommand{\newblock}{\relax}
\providecommand{\bibinfo}[2]{#2}
\providecommand{\BIBentrySTDinterwordspacing}{\spaceskip=0pt\relax}
\providecommand{\BIBentryALTinterwordstretchfactor}{4}
\providecommand{\BIBentryALTinterwordspacing}{\spaceskip=\fontdimen2\font plus
\BIBentryALTinterwordstretchfactor\fontdimen3\font minus
  \fontdimen4\font\relax}
\providecommand{\BIBforeignlanguage}[2]{{%
\expandafter\ifx\csname l@#1\endcsname\relax
\typeout{** WARNING: IEEEtran.bst: No hyphenation pattern has been}%
\typeout{** loaded for the language `#1'. Using the pattern for}%
\typeout{** the default language instead.}%
\else
\language=\csname l@#1\endcsname
\fi
#2}}
\providecommand{\BIBdecl}{\relax}
\BIBdecl

\bibitem{abeel2009robust}
T.~Abeel, T.~Helleputte, Y.~Van~de Peer, P.~Dupont, and Y.~Saeys, ``Robust
  biomarker identification for cancer diagnosis with ensemble feature selection
  methods,'' \emph{Bioinformatics}, vol.~26, no.~3, pp. 392--398, 2009.

\bibitem{he2010stable}
Z.~He and W.~Yu, ``Stable feature selection for biomarker discovery,''
  \emph{Computational biology and chemistry}, vol.~34, no.~4, pp. 215--225,
  2010.

\bibitem{donoho2006compressed}
D.~L. Donoho \emph{et~al.}, ``Compressed sensing,'' \emph{IEEE Transactions on
  information theory}, vol.~52, no.~4, pp. 1289--1306, 2006.

\bibitem{wright09}
S.~J. Wright, R.~D. Nowak, and M.~A. Figueiredo, ``Sparse reconstruction by
  separable approximation,'' \emph{IEEE Transactions on signal processing},
  vol.~57, no.~7, pp. 2479--2493, 2009.

\bibitem{Alv2016}
J.~M. Alvarez and M.~Salzmann, ``Learning the number of neurons in deep
  networks,'' in \emph{Advances in Neural Information Processing Systems},
  2016, pp. 2270--2278.

\bibitem{Han2015-2}
S.~Han, J.~Pool, J.~Tran, and W.~Dally, ``Learning both weights and connections
  for efficient neural network,'' in \emph{Advances in neural information
  processing systems}, 2015, pp. 1135--1143.

\bibitem{Tar2018}
E.~Tartaglione, S.~Leps{\o}y, A.~Fiandrotti, and G.~Francini, ``Learning sparse
  neural networks via sensitivity-driven regularization,'' in \emph{Advances in
  Neural Information Processing Systems}, 2018, pp. 3878--3888.

\bibitem{Gom2019}
A.~N. Gomez, I.~Zhang, K.~Swersky, Y.~Gal, and G.~E. Hinton, ``Learning sparse
  networks using targeted dropout,'' \emph{arXiv :1905.13678}, 2019.

\bibitem{Osw2016}
U.~Oswal, C.~Cox, M.~Lambon-Ralph, T.~Rogers, and R.~Nowak, ``Representational
  similarity learning with application to brain networks,'' in
  \emph{International Conference on Machine Learning}, 2016, pp. 1041--1049.

\bibitem{dropout1}
N.~Srivastava, G.~Hinton, A.~Krizhevsky, I.~Sutskever, and R.~Salakhutdinov,
  ``Dropout: a simple way to prevent neural networks from overfitting,''
  \emph{The journal of machine learning research}, vol.~15, no.~1, pp.
  1929--1958, 2014.

\bibitem{dropout2}
J.~Cavazza, P.~Morerio, B.~Haeffele, C.~Lane, V.~Murino, and R.~Vidal,
  ``Dropout as a low-rank regularizer for matrix factorization,'' in
  \emph{International Conference on Artificial Intelligence and Statistics
  (AISTATS)}, 2018, pp. 435--444.

\bibitem{tRS}
R.~Tibshirani, ``Regression shrinkage and selection via the lasso,''
  \emph{Journal of the Royal Statistical Society. Series B (Methodological)},
  pp. 267--288, 1996.

\bibitem{sparse}
T.~Hastie, R.~Tibshirani, and M.~Wainwright, ``Statistcal learning with
  sparsity: The lasso and generalizations,'' \emph{CRC Press}, 2015.

\bibitem{condat}
L.~Condat, ``Fast projection onto the simplex and the l1 ball,''
  \emph{Mathematical Programming Series A}, vol. 158, no.~1, pp. 575--585,
  2016.

\bibitem{Perez19}
G.~Perez, M.~Barlaud, L.~Fillatre, and J.-C. R{\'e}gin, ``A filtered
  bucket-clustering method for projection onto the simplex and the
  $\ell_1$-ball,'' \emph{Mathematical Programming}, May 2019.

\bibitem{Yua}
M.~Yuan and Y.~Lin, ``Model selection and estimation in regression with grouped
  variables,'' \emph{Journal of the Royal Statistical Society: Series B
  (Statistical Methodology)}, vol.~68, no.~1, pp. 49--67, 2006.

\bibitem{Hua2018}
Z.~Huang and N.~Wang, ``Data-driven sparse structure selection for deep neural
  networks,'' in \emph{Proceedings of the European Conference on Computer
  Vision (ECCV)}, 2018, pp. 304--320.

\bibitem{Yoo2017}
J.~Yoon and S.~J. Hwang, ``Combined group and exclusive sparsity for deep
  neural networks,'' in \emph{Proceedings of the 34th International Conference
  on Machine Learning-Volume 70}.\hskip 1em plus 0.5em minus 0.4em\relax JMLR.
  org, 2017, pp. 3958--3966.

\bibitem{Sca2017}
S.~Scardapane, D.~Comminiello, A.~Hussain, and A.~Uncini, ``Group sparse
  regularization for deep neural networks,'' \emph{Neurocomputing}, vol. 241,
  pp. 81--89, 2017.

\bibitem{BCD}
N.~Simon, J.~Friedman, T.~Hastie, and R.~Tibshirani, ``A sparse-group lasso,''
  \emph{Journal of Computational and Graphical Statistics}, vol.~22, no.~2, pp.
  231--245, 2013.

\bibitem{Fastlasso}
I.~Yasutoshi, F.~Yasuhiro, and K.~Hisashi, ``Fast sparse group lasso,'' in
  \emph{Advances in Neural Information Processing Systems}, vol.~32.\hskip 1em
  plus 0.5em minus 0.4em\relax Curran Associates, Inc., 2019.

\bibitem{BG20}
M.~Barlaud and F.~Guyard, ``Learning sparse deep neural networks using
  efficient structured projections on convex constraints for green ai,''
  \emph{International Conference on Pattern Recognition, Milan}, pp.
  1566--1573, 2020.

\bibitem{quattoni2009efficient}
A.~Quattoni, X.~Carreras, M.~Collins, and T.~Darrell, ``An efficient projection
  for $\ell_{1,\infty}$ regularization,'' in \emph{Proceedings of the 26th
  Annual International Conference on Machine Learning}, 2009, pp. 857--864.

\bibitem{Rakoto}
A.~Rakotomamonjy, R.~Flamary, G.~Gasso, and S.~Canu, ``lp-lq penalty for sparse
  linear and sparse multiple kernel multitask learning,'' \emph{IEEE
  Transactions on Neural Networks}, vol.~22, p. 307–1320, 2011.

\bibitem{bejar2021fastest}
B.~Bejar, I.~Dokmani{\'c}, and R.~Vidal, ``The fastest $\ell_{1,\infty}$ prox
  in the {W}est,'' \emph{IEEE transactions on pattern analysis and machine
  intelligence}, vol.~44, no.~7, pp. 3858--3869, 2021.

\bibitem{chau2019efficient}
G.~Chau, B.~Wohlberg, and P.~Rodriguez, ``Efficient projection onto the
  $\ell_{1,\infty}$ mixed-norm ball using a newton root search method,''
  \emph{SIAM Journal on Imaging Sciences}, vol.~12, no.~1, pp. 604--623, 2019.

\bibitem{chu2020semismooth}
D.~Chu, C.~Zhang, S.~Sun, and Q.~Tao, ``Semismooth newton algorithm for
  efficient projections onto $\ell_{1,\infty}$-norm ball,'' in
  \emph{International Conference on Machine Learning}, 2020, pp. 1974--1983.

\bibitem{bilevel2}
A.~Sinha, P.~Malo, and K.~Deb, ``A review on bilevel optimization: From
  classical to evolutionary approaches and applications,'' \emph{IEEE
  Transactions on Evolutionary Computation}, vol.~22, no.~2, pp. 276--295,
  2018.

\bibitem{bilevel3}
K.~Bennett, J.~Hu, X.~Ji, G.~Kunapuli, and J.-S. Pang, ``Model selection via
  bilevel optimization,'' \emph{IEEE International Conference on Neural
  Networks - Conference Proceedings}, 2006.

\bibitem{Perez-2023}
G.~Perez, L.~Condat, and M.~Barlaud, ``Near-linear time projection onto the
  l1,infty ball application to sparse autoencoders.'' \emph{arXiv: 2307.09836},
  2023.

\bibitem{Twitter}
L.~Theis, W.~Shi, A.~Cunningham, and F.~Huszár, ``Lossy image compression with
  compressive autoencoders,'' \emph{ICLR Conference Toulon}, 2017.

\bibitem{Mentzer}
F.~Mentzer, G.~Toderici, M.~Tschannen, and E.~Agustsson, ``High-fidelity
  generative image compression,'' \emph{NEURIPS}, 2020.

\bibitem{JPEGAI}
J.~Ascenso, E.~Alshina, and T.~Ebrahimi, ``The jpeg ai standard: Providing
  efficient human and machine visual data consumption,'' \emph{IEEE
  MultiMedia}, vol.~30, no.~1, pp. 100--111, 2023.

\bibitem{deep}
I.~Goodfellow, Y.~Bengio, and A.~Courville, \emph{Deep learning}.\hskip 1em
  plus 0.5em minus 0.4em\relax MIT press, 2016, vol.~1.

\bibitem{VAE}
D.~Kingma and M.~Welling, ``Auto-encoding variational bayes,''
  \emph{International Conference on Learning Representation}, 2014.

\bibitem{semisupervised}
D.~P. Kingma, S.~Mohamed, D.~Jimenez~Rezende, and M.~Welling, ``Semi-supervised
  learning with deep generative models,'' \emph{Advances in neural information
  processing systems}, vol.~27, 2014.

\bibitem{Rochelle}
J.~Snoek, R.~Adams, and H.~Larochelle, ``On nonparametric guidance for learning
  autoencoder representations,'' in \emph{Artificial Intelligence and
  Statistics}.\hskip 1em plus 0.5em minus 0.4em\relax PMLR, 2012, pp.
  1073--1080.

\bibitem{LESAE}
L.~Le, A.~Patterson, and M.~White, ``Supervised autoencoders: Improving
  generalization performance with unsupervised regularizers,'' \emph{Advances
  in Neural Information Processing Systems}, 2018.

\bibitem{ICASSP}
M.~Barlaud and F.~Guyard, ``Learning a sparse generative non-parametric
  supervised autoencoder,'' \emph{Proceedings of the International Conference
  on Acoustics, Speech and Signal Processing, Toronto, Canada}, June 2021.

\bibitem{Huber}
P.~J. Huber, \emph{Robust statistics}.\hskip 1em plus 0.5em minus 0.4em\relax
  Wiley, New York, 1981.

\bibitem{BBCF}
M.~Barlaud, W.~Belhajali, P.~Combettes, and L.~Fillatre, ``Classification and
  regression using an outer approximation projection-gradient method,''
  vol.~65, no.~17, 2017, pp. 4635--4643.

\bibitem{hrtzER}
T.~Hastie, S.~Rosset, R.~Tibshirani, and J.~Zhu, ``The entire regularization
  path for the support vector machine,'' \emph{Journal of Machine Learning
  Research}, vol.~5, pp. 1391--1415, 2004.

\bibitem{fht}
J.~Friedman, T.~Hastie, and R.~Tibshirani, ``Regularization path for
  generalized linear models via coordinate descent,'' \emph{Journal of
  Statistical Software}, vol.~33, pp. 1--122, 2010.

\bibitem{myCA}
J.~Mairal and B.~Yu, ``Complexity analysis of the lasso regularization path,''
  in \emph{Proceedings of the 29th International Conference on Machine Learning
  (ICML-12)}, 2012, pp. 353--360.

\bibitem{Lottery}
J.~Frankle and M.~Carbin, ``The lottery ticket hypothesis: Finding sparse,
  trainable neural networks,'' \emph{arXiv preprint arXiv:1803.03635}, 2018.

\bibitem{double}
H.~Zhou, J.~Lan, R.~Liu, and J.~Yosinski, ``Deconstructing lottery tickets:
  Zeros, signs, and the supermask,'' in \emph{Advances in Neural Information
  Processing Systems 32}, 2019, pp. 3597--3607.

\bibitem{Adam}
D.~Kingma and J.~Ba, ``a method for stochastic optimization.''
  \emph{International Conference on Learning Representations}, pp. 1--13, 2015.

\bibitem{Lung}
E.~{Math\'e \emph{et al.}}, ``Noninvasive urinary metabolomic profiling
  identifies diagnostic and prognostic markers in lung cancer,'' \emph{Cancer
  research}, vol.~74, no.~12, p. 3259—3270, June 2014.

\end{thebibliography}
\bibliographystyle{IEEEtran}

\newpage
\appendix
\onecolumn

\section{Iterative form of the multi-level projection}
In the paper we provided a recursive form of the tri-level algorithm and 
an iterative form for the multi-level as they are easier to understand.
We provided here their iterative version.

\begin{algorithm}[h]
   \caption{Tri-level iterative projection $\ell_{1,\infty,\infty}$.}\label{algo:linfimageprojIter}
\begin{algorithmic}[1]
\STATE \textbf{Input:} $\mathcal{Y} \in \RR^{c,n,m},\eta$
\STATE{$u^{3} \gets P^{1}_\eta(((v_{\infty,\infty})_t, \forall  t \in m))$} 
\FOR{$ t \in <m>$}
    \STATE $U^{2}_t \gets P^{\infty}_{u^{3}_t}((V_{{\infty}})_t)$ 
\ENDFOR
\FOR{$ t \in <n,m>$}
    \STATE $\mathcal{U}^{1}_t \gets P^{\infty}_{U^{2}_t}(\mathcal{Y}_{t})$ 
\ENDFOR
\STATE \textbf{Output:} $\mathcal{U}^{1}$
\end{algorithmic}
\end{algorithm}

\begin{algorithm}[h]
   \caption{Multi-level iterative projection.}\label{algo:linfmultiprojIter}
\begin{algorithmic}[1] 
\STATE \textbf{Input:} $\mathcal{Y},\nu,\eta$
\STATE{$\mathcal{U}^{|\nu|} \gets P^{\nu_{|\nu|}}_\eta(((\mathcal{V}_{{\nu_1,\dots,\nu_{|\nu|-1}}})_t, \forall  t \in T_{|T|-|\nu_{|\nu|}|:|T|}))$}  
\FOR{$ \nu_i \in (\nu_{|\nu|-1},\dots,\nu_1)$}
    \FOR{$ t \in <T_{|T|-\sum_{j=i+1}^{|\nu|}|\nu_j|:|T|}>$}
        \STATE $\mathcal{U}^{i}_t \gets P^{\nu_i}_{\mathcal{U}^{i-1}_t}((\mathcal{V}_{{\nu_1,\dots,\nu_{i-1}}})_t)$ 
    \ENDFOR
\ENDFOR
\STATE \textbf{Output:} $\mathcal{U}^{1}$
\end{algorithmic}
\end{algorithm}
\section{Proofs of propositions }
\textbf{Proof of proposition 6.3.} For any norm $\calN$ and radius $\eta$, let $\nu=(\calN)$. 
Then, since $|\nu|=1$, we have $MP^{\nu}_\eta(Y)=P^{\calN}_\eta(Y)$.

\textbf{Proof of proposition 6.4} The complexity of the recursive algorithm~\ref{algo:linfmultiproj} is split in two parts:
\textbf{1) Aggregations}) 
At line \ref{algo:linfmultiproj:aggreg} tensor $\mathcal{Y}$ is aggregated using the first norm presents in $\nu$.
This will be done by each call of the $MP^{\nu}_\eta(\mathcal{Y})$ until $\nu$ is a singleton.
For a given norm $\nu_i$, the current aggregated tensor is split into independent sub-parts and the norm $\nu_i$ of each of these sub-parts is processed.
For the two-level case, the aggregation is made of many independent processes that can run in parallel.
The time complexity of aggregation for norm $\nu_i$ is the sum of the time complexity of processing the norm $\nu_i$ for each sub-part $t$ $O(\sum_t f_i(|\nu_i|))$.
With infinite parallel processing power, the time complexity can be reduced to $O(f_i(|\nu_i|))$, as each sub-part is independent.
\textbf{2) Projections}) 
Once the aggregated tensor is processed, it is projected.
The result of each of these projections is stored into $t$.
The most inner one, in the deepest call of the recursive processing, is the usual projection of the aggregated tensor onto the norm $\nu_{|\nu|}$ of radius $\eta$.
The time complexity of this line is $O(f_p(|\nu_{|\nu|}|))$ which is the time to project onto the norm $\nu_{|\nu|}$.
Then, the algorithm uses the result of the previous projection ($t^2$) to project its current aggregation of the tensor,
given as input argument in the recursive calls.
This is done in line \ref{algo:linfmultiproj:project} where the result of the previous projection ($t^{2}_t$, a value) is used to compute the current projection ($t^{1}_t$, a tensor).
Again, the loop on the indices $t$, at line \ref{algo:linfmultiproj:loopt},
is made of independent sub-parts that can be run in parallel.
This implies that with infinite parallel processing power, the time complexity of the projections is $O(\sum_i f_{p,i}(|\nu_i|))$.
Compared to the general case, the parallel computation may provide an exponential speedup.
An iterative implementation of the algorithm is given in Algorithm~\ref{algo:linfmultiprojIter} to explicitly show the different computations.

\end{document}